# Class-Conditional Conformal Triage for ICAO Face-Image Compliance under Class Imbalance

**Mustafa Vedat YILDIRIM and Inna SKARGA-BANDUROVA**
School of Engineering, Computing and Mathematics, Oxford Brookes University, Headington Road, Oxford OX3 0BP, United Kingdom
e-mail: 19341298@brookes.ac.uk, iskarga-bandurova@brookes.ac.uk

ABSTRACT Automated ICAO/ISO face-image compliance analysers provide criterion-level pass/fail decisions, yet these verdicts do not quantify uncertainty when deciding whether to accept, reject, or retake a photograph. We investigate conformal prediction as a triage layer for BioGaze, evaluating 14 compliance criteria on the synthetic TONO benchmark. Without retraining the analyser, we fit per-criterion logistic classifiers and compare marginal and label-conditional (Mondrian) conformal prediction with conventional abstention rules using subject-disjoint splits. Marginal conformal prediction achieved approximately 0.90 overall coverage but substantially under-covered violations, with a ten-partition mean of 0.37 for one criterion. The corresponding Mondrian means ranged from 0.89 to 0.92 across criteria. However, most of the reduction in criterion-level false positives came from the logistic layer (0.145 to 0.048), while tuned abstention rules achieved lower false-positive rates than Mondrian at comparable retake rates. Bonferroni correction for joint coverage across 14 criteria required retaking nearly all images. These findings show that class-conditional conformal prediction can achieve near-target empirical coverage for minority violation classes and reveal trade-offs between statistical coverage, classification performance, and retake requirements in multi-criterion face-image compliance assessment.

## I. INTRODUCTION

Face images used in passports and other identity documents must meet the photographic requirements of ICAO Doc 9303 [1] and ISO/IEC 39794-5 [2]. These include a frontal pose, neutral expression, open eyes, uniform lighting and background, absence of sunglasses or headwear, and adequate sharpness, exposure, and colour. Numerous automated systems now assess compliance with these requirements during enrolment, reducing the need for manual inspection. However, most of them provide a simple pass/fail decision for each criterion without indicating how reliable that decision is. This creates two types of errors with different consequences: incorrectly rejecting a compliant photograph requires an unnecessary retake, while accepting a non-compliant photograph may compromise its suitability for an identity document.

The problem becomes more challenging when violations are less frequent than compliant cases for individual criteria. A system may achieve high overall accuracy while still missing a substantial proportion of violations. Moreover, a single photograph must satisfy multiple requirements, and uncertainty in any of these assessments can affect the final decision. Rather than forcing every image into an accept or reject category, a more useful approach is to distinguish between clear compliance, clear violations, and cases that require another photograph. This motivates a three-way decision process: ACCEPT, REJECT, or RETAKE.

In this study, we investigate whether conformal prediction can provide a principled basis for such decisions. We build on BioGaze [3], an open-source analyser that evaluates individual ICAO/ISO compliance criteria. Without modifying or retraining BioGaze, we use its outputs to train lightweight classifiers and apply split conformal prediction to produce sets of plausible labels for each criterion. We compare standard marginal conformal prediction with its label-conditional (Mondrian) variant, which is designed to control coverage separately for compliant and non-compliant cases under exchangeability. This distinction matters because high overall coverage does not necessarily imply reliable coverage of the less frequent violation class.

We evaluate the approach on 14 compliance criteria using the synthetic TONO benchmark, separating subjects across training, calibration, and testing to prevent identity leakage. Our analysis examines how marginal and class-conditional coverage differ, how conformal prediction compares with conventional abstention strategies, and what happens when decisions from multiple criteria are combined into a single image-level assessment. By considering prediction coverage alongside classification errors and retake requirements, the study provides a systematic evaluation of the benefits and practical costs of uncertainty-aware face-image compliance assessment.

This study makes four **contributions:**

- We extend the rule-based ICAO BioGaze analyser with lightweight per-criterion classifiers and conformal prediction sets, enabling ACCEPT, REJECT, and RETAKE decisions without modifying the underlying analyser.
- We show that marginal conformal prediction can substantially under-cover the minority violation class despite achieving its overall coverage target, consistent with its known behaviour on minority classes (Section

II). Across ten subject-disjoint partitions, Mondrian conformal prediction brings mean violation-class coverage close to the nominal 0.90 level across all 14 criteria.
- Through feature ablation and comparison of seven triage strategies, we distinguish improvements in classification performance from those in prediction-set coverage. Most of the reduction in false positives comes from the logistic classifiers, while conventional abstention rules achieve competitive or better empirical operating points. Mondrian's distinctive value lies in its class-conditional coverage property under exchangeability.
- We examine how uncertainty across 14 compliance criteria affects image-level decisions and joint coverage. We show that Bonferroni correction can recover near-target empirical joint coverage, but only at the cost of retaking nearly all images, revealing a practical limitation of Bonferroni-corrected simultaneous coverage in multi-criterion compliance assessment.

The remainder of this paper is organised as follows. Section II reviews conformal prediction, selective classification and automated ICAO compliance. Section III describes the method, the subject-disjoint protocol and the seven triage strategies. Section IV presents the results. Section V discusses operational considerations, external validity and limitations. Finally, Section VI concludes.

## II. RELATED WORK

We compare the least ambiguous set-valued classifier (LAC) nonconformity score of Sadinle, Lei & Wasserman [4] with the Adaptive Prediction Sets of Romano, Sesia & Candès [5]. Conformal prediction provides distribution-free finite-sample coverage for any base predictor, assuming exchangeability [6], [7], [8]. Marginal forms of conformal prediction can yield low coverage for a minority class. Label-conditional (Mondrian) conformal prediction is calibrated for every single class and provides a coverage guarantee for each class [6], [9]; if there are very few calibration examples available for certain classes, then clustered conformal prediction can be used to combine similar classes [10]. Our study considers the binary version of this concept applied to multiple decision-making problems, specifically 14 coupled criteria. This creates a joint coverage question, essentially a multiple-comparisons question, that would not arise in studies of individual decisions.

The ACCEPT, REJECT, or RETAKE decision process relates to selective prediction and the reject option [11], [12], [13], [14]. The asymmetric costs of a miss and a retake relate to conformal risk control [15], [16], which could certify a cost-weighted risk directly; we do not use it here. Automated ICAO/ISO compliance verification has primarily focused on point classification. Ferrara et al. [17] developed compliance checking and the BioLab-ICAO benchmark.

Research on face image quality assessment [18], [19] has extensively investigated methods such as the open-source OFIQ tool [20], with researchers assigning scalar quality scores or verdicts for specific criteria. A recent framework in this field is BioGaze by Elatfi et al. [3], which combines traditional computer vision techniques with deep learning models to verify compliance with ISO/IEC 39794-5 photographic requirements.

We incorporate BioGaze as an analysis tool. Its companion datasets, ONOT [21] and TONO [22], are synthetically generated. Di Domenico et al. [23] developed a zero-shot model using CLIP-IQA and natural language prompts. Gonçalves et al. [24] developed the DFIC dataset: 58,633 images of 1,016 subjects, including 16,644 artificially generated non-compliant images, annotated against 26 ICAO requirements, with a demographically balanced training partition. The authors compared BioGaze, OFIQ and ICAONet [25] on this dataset and assessed how well these models generalise to TONO. All these methods produce either definitive verdicts or scalar scores for each requirement but do not offer a distribution-free coverage guarantee for their decisions. Additionally, they do not distinguish between a confident rejection and situations where the subject should retake the photo. Finally, none report coverage on the violation class, which is exactly the gap we aim to fill. Since our layer is independent of the underlying analyser, it may be used with any of these systems. Furthermore, because DFIC contains a truly compliant class, it is an ideal choice for continuing research on real-world data.

Recent research has also examined how conformal prediction supports decisions with an abstention option. Lee et al. [26] developed hierarchical conformal prediction (HCP) for grouped data, which is relevant when multiple correlated photographs belong to the same subject. Hallberg Szabadváry et al. [27] formalised binary classification with a reject option, distinguishing errors in confident predictions from the rejection rate. In a related biometric setting, Fatima and Ahmed [28] applied conformal prediction to coverage-controlled face and pedestrian retrieval, where the objective is to retain relevant identities in a candidate set rather than assess photographic compliance. Building on these complementary research directions, our study examines how marginal and class-conditional prediction sets behave across 14 ICAO compliance criteria. By comparing conformal and conventional abstention strategies and analysing their combined image-level decisions, we characterise the trade-offs between violation-class coverage, classification performance, and retake requirements in multi-criterion compliance assessment.

## III. METHOD

The proposed pipeline consists of three stages: criterion-level scoring, conformal calibration, and decision-making. First, BioGaze extracts criterion-level decisions and auxiliary measurements from each image. These outputs are used to train a separate logistic classifier for each of the 14 compliance criteria. The resulting scores are converted into prediction sets using split conformal prediction [29]. Finally, the prediction sets are combined to produce ACCEPT, REJECT, or RETAKE decisions.

This design separates three distinct functions: the classifier estimates the likelihood of a violation, conformal calibration determines which labels remain plausible at a specified nominal coverage level, and the decision rule determines whether the evidence supports a definitive action. Fig. 1 illustrates the complete pipeline.

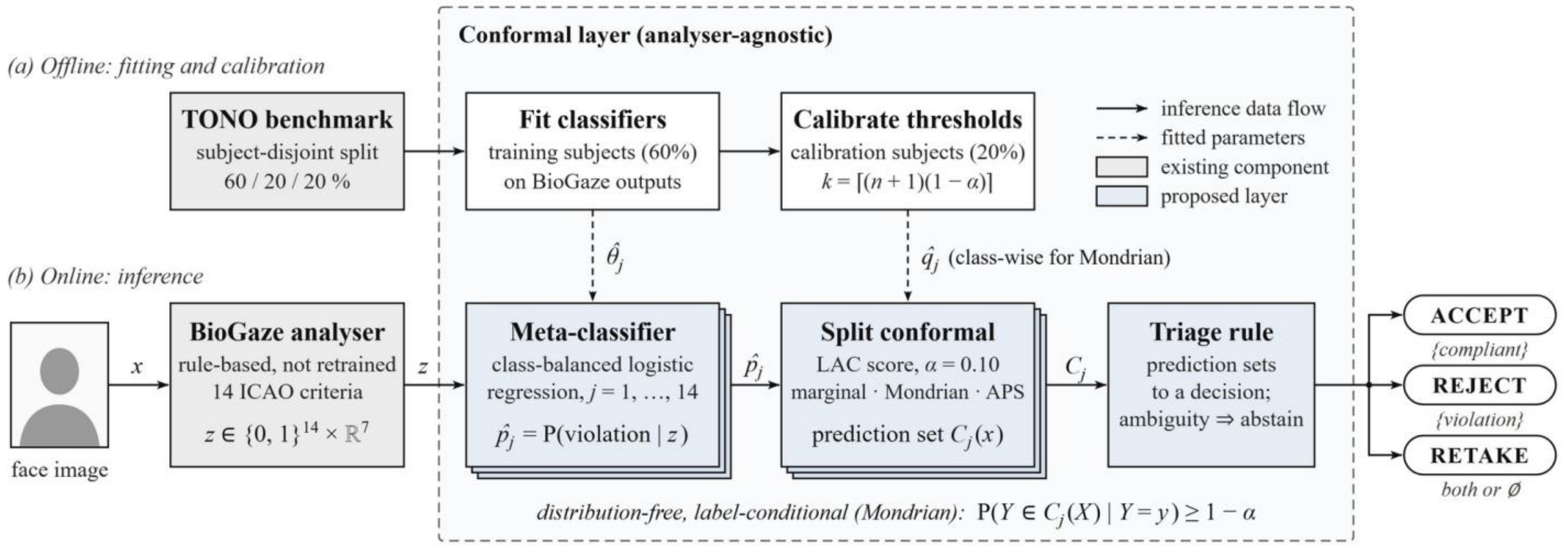

**Figure 1.** Conformal layer on a rule-based ICAO analyser: BioGaze features, a classifier for each criterion, split conformal prediction sets, and an accept, reject or retake decision. The layer does not depend on the analyser.

### *A. FEATURES AND PER-CRITERION META-CLASSIFIER*

BioGaze produces 14 binary compliance decisions for each image, together with seven auxiliary measurements and indicators, including inter-eye distance, head pose angles, eye and mouth openness, and the presence of glasses. These outputs form the input to our criterion-level classifiers.

For each criterion $j$, we define a binary ground-truth label, with 1 indicating a violation and 0 indicating compliance. We train a separate class-balanced logistic regression model to distinguish these classes using the BioGaze outputs.

Let $z(x)$ denote the vector containing the 14 binary decisions and seven auxiliary features for image $x$. The estimated violation score for criterion $j$ is

$$\hat{p}_j(1 \mid x) = \sigma(w_j^T z(x) + b_j), \quad (1)$$

where σ is the logistic sigmoid function, $w_j$ and $b_j$ are the learned parameters, and $\hat{p}_j(0 \mid x) = 1 - \hat{p}_j(1 \mid x)$.

The classifiers are fitted using the training data only. Their conformal thresholds are then computed on a separate set of images, while the original BioGaze analyser remains fixed. Because class weighting can affect probability calibration, the logistic outputs are treated as predictive scores rather than guaranteed calibrated probabilities. Conformal prediction does not require perfect probability calibration to achieve its theoretical coverage property under the appropriate exchangeability assumptions.

The primary model uses all 21 BioGaze features, allowing each criterion classifier to exploit information from other compliance checks. To examine the effect of these cross-criterion dependencies, we compare three feature configurations: the complete feature set, the criterion's own binary decision together with the auxiliary features, and the auxiliary features alone. This comparison is particularly relevant to TONO, where each in-scope image is constructed to violate a single criterion. Information from other compliance checks may therefore help identify the target violation, but the resulting classification improvements may not generalise to images with multiple simultaneous violations. The feature ablation in Section IV.A examines this dependence.

### *B. DATASET AND SUBJECT-DISJOINT EVALUATION*

We evaluate the proposed approach using the synthetic TONO benchmark [22]. The TONO subject identifier is extracted from every filename. A majority (247) of the 275 identities were found in multiple violation folders. If identities were divided by violation type, the same identity would appear in both calibration and test, inflating coverage.

Therefore, we divided identities, not images, into three groups: train, calibration, and test, using a 60/20/20 ratio. These three groups had no identity overlap (verified). One fixed partition (partition 1) was used for all primary statistical analyses. We report coverage for each criterion, with 95% confidence intervals, using a cluster bootstrap [30]. A cluster bootstrap is a resampling technique in which we resample the 55 test subjects with replacement, treating each subject's images as a unit. By design, this preserves the correlation among data points from the same subject. Using multiple partitions enabled a robustness analysis, and we report the results as the mean, standard deviation, and range across ten partitions (see Section IV.B). The comparison of methods, aggregated across five partitions, is summarised by means and standard deviations (see Section IV.C).

Each subject has many correlated images (median of 14-15 images per subject in each split). Therefore, the analysis assumes that calibration and test scores at the image level are exchangeable, and this exchangeability is a requirement of the analysis rather than a result of the split. The guarantee holds at the image level under the exchangeability assumption, and Section IV.G examines the sensitivity of the results to within-subject dependence.

**Scope of the label space.** The TONO "zoom" folder contains images where the head is too large for the ICAO framing envelope. BioGaze does not check for this head-size constraint. As a result, none of these images have been given labels within our scope and will not be included in any of our coverage or trade-off analyses (Sections IV.A to IV.C and IV.E to IV.G). We report them only in the image counts in Section IV.D, where we list them separately as out of scope. By construction, every image within our scope violates only one criterion.

### *C. SPLIT CONFORMAL: MARGINAL, MONDRIAN, AND APS*

We use split conformal prediction to convert the logistic classifier scores into sets of plausible labels. For each criterion, the least ambiguous set-valued classifier (LAC) method [4] assigns a nonconformity score to an image $x$ and candidate label $y$:

$$s(x, y) = 1 - \hat{p}(y \mid x), \quad (2)$$

where $\hat{p}(y \mid x)$ is the estimated probability of that label. Lower nonconformity scores indicate stronger support for the candidate label.

For a calibration set containing $n$ scores, the conformal threshold $\hat{q}$ is defined as the $k$-th smallest calibration score, where

$$k = \lceil (n + 1)(1 - \alpha) \rceil. \quad (3)$$

The parameter $\alpha$ specifies the nominal miscoverage level. If k exceeds n, the threshold is set to infinity, and every label calibrated with that threshold enters the prediction set. At $\alpha$ = 0.10, this occurs when fewer than nine calibration scores are available.

For a new image, the prediction set contains every label whose nonconformity score does not exceed the calibration threshold:

$$C(x) = \{y : 1 - \hat{p}(y \mid x) \leq \hat{q}\}. \quad (4)$$

We evaluate two forms of LAC calibration. Marginal conformal prediction uses a single calibration threshold for both classes within each criterion. This controls overall prediction-set coverage under exchangeability, but it does not guarantee the same coverage separately for compliant and violation cases.

Label-conditional (Mondrian) conformal prediction instead calibrates separate thresholds for the two classes. For each criterion $j$ and class $y$, its threshold is calculated from the calibration examples belonging to that class. Under the required exchangeability assumptions, this construction provides the class-conditional coverage property

$$P(Y \in C_j(X) \mid Y = y) \geq 1 - \alpha, \quad y \in \{0, 1\}. \quad (5)$$

This guarantee is marginal over the relevant calibration and test draws; it does not imply that every individual test partition reaches the nominal level. In particular, Mondrian calibration targets coverage of the violation class directly rather than allowing the more frequent compliant class to dominate the overall coverage estimate.

**Interpretation of validity.** For a given criterion, conformal miscoverage is the event that the true label is excluded from its prediction set. This differs from confident-violation recall, since an ambiguous set contains the true violation label but produces RETAKE. For the singleton-compliant decision, the probability of a false-compliant action on a true violation never exceeds the violation-class miscoverage probability; under Mondrian calibration and the stated exchangeability assumptions, this probability is at most α. A simultaneous 14-label guarantee and an operational safety guarantee for all image-level actions are different targets.

In addition to LAC, we evaluate randomised Adaptive Prediction Sets (APS) [5]. We developed the layer from scratch to fully control its implementation and verified it against the MAPIE library [31] in the marginal case: MAPIE returned identical prediction sets on every test image, for all 14 criteria and ten partitions. This comparison verifies agreement with the reference implementation for marginal LAC.

### D. TRIAGE STRATEGIES AND BASELINES

We compare seven decision strategies: the original BioGaze binary decisions, a logistic classification without abstention, Chow's confidence-based rejection rule [11], an asymmetric two-threshold rule, marginal LAC, Mondrian conformal prediction, and randomised APS across the same five partitions. The first two methods provide definitive binary decisions and serve as non-abstaining baselines. The remaining methods allow uncertain cases to be assigned to RETAKE rather than forcing a compliant or violation decision.

**Conventional abstention rules.** Chow's confidence rejection rule assigns RETAKE when the largest predicted class score, max(p, 1 − p), falls below a confidence threshold τ. Otherwise, it selects the class with the higher score. The threshold is selected on the calibration set to obtain an abstention rate as close as possible to a specified target.

The second is an asymmetric rule with two thresholds. This rule decides compliant when p is less than or equal to $t_{lo}$, violation when p is greater than or equal to $t_{hi}$ and abstains when p falls between $t_{lo}$ and $t_{hi}$. These two thresholds minimise a cost of 10 per missed violation plus 1 per false flag for decided cases. All tuned quantities are calibrated on the calibration split only. τ is chosen to obtain a calibration abstention rate as close as possible to a target design rate of 0.10. Similarly, ($t_{lo}$, $t_{hi}$) are chosen to yield a calibration abstention rate within 0.02 of the same target. For several criteria, no pair of thresholds on the grid meets the specified tolerance, so we apply the rule with a fixed band (i.e., compliant if p ≤ 0.4, violation if p ≥ 0.6) without tuning the cost. This occurred in 27 of 70 criterion fits at the 0.10 target, and therefore the pooled calibration abstention rate for this rule was 0.07. The predetermined target rate of 0.10 serves as a reference point for comparing the RETAKE rates across methods.

**Conformal decision rules.** For LAC, Mondrian, and APS, the nominal miscoverage level is set to α = 0.10 in the principal comparison. Unlike the conventional abstention target, α controls prediction-set coverage rather than the proportion of cases sent to RETAKE. The resulting abstention rates are therefore measured empirically and are not expected to match those of the conventional rules exactly. On the test splits, Mondrian at α = 0.10 sends 10.4% of criterion decisions to RETAKE on average (Table 3), close to the 0.10 design target. The 0.10 target was not derived from test data.

For each criterion, conformal prediction sets are translated into operational actions as follows: {violation} → REJECT; {compliant} → ACCEPT; {compliant, violation} or an empty set → RETAKE. An image is rejected if at least one criterion produces a confident violation. It is accepted only when all 14 criteria produce confident compliant decisions. All remaining combinations result in RETAKE. Thus, a confident violation takes precedence over uncertainty in other criteria, while acceptance requires unanimous confident compliance.

**Comparison and evaluation.** All logistic classifiers are fitted on the training split only, and all threshold selection is performed using calibration data. The test split is used only to measure performance. Section IV.C reports abstention rates for the abstention rules at targets 0.05 and 0.15 to avoid concluding solely on a single target. As a calibrated baseline for the view of recall against false-positive rate, we fit isotonic regression [32] on the calibration split and sweep through its threshold.

We evaluate each strategy using criterion-level false-positive rate, violation recall, RETAKE rate, and selective error. False-positive rate measures confident violation predictions among truly compliant criterion cases; abstained cases remain in its denominator. Recall over all violations counts RETAKE as a case not confidently detected, while recall among decided cases excludes abstentions. Selective error measures classification errors among cases receiving definitive decisions. These metrics distinguish improvements in classification performance from reductions achieved by withholding uncertain decisions.

Conformal prediction-set coverage is reported separately from these operational metrics. This allows us to compare conventional abstention and conformal strategies while preserving the distinction between empirical classification performance and the statistical coverage properties of conformal prediction.

## IV. RESULTS

### A. THE GUARANTEE DOES NOT DEPEND ON THE FEATURES; THE ACCURACY IS A DATASET ARTEFACT

The relationship between the criteria for BioGaze decisions and the fact that each TONO image has exactly one violation is such that if a criterion classifier flags an additional criterion, the image will most likely comply. Table 1 describes this relationship. Using the complete feature set in the logistic layer reduces BioGaze's false positive rate by 67%; however, when only the criterion's own decision and the continuous metrics are used, the reduction in false positives drops to 26%; and when only the continuous metrics are used, the false positive rate increases beyond that of the raw analyser. Conformal coverage is fairly consistent across all three feature sets, as predicted by the distribution-free theory.

**Table 1.** Feature ablation (five seeds with disjoint subjects). Empirical coverage is similar across feature sets; logistic accuracy is not, and with weak features coverage is kept only by producing more ambiguous sets.

| Feature set | Logistic FPR | False-flag redn. | Mond. viol. cov. | Mond. set size | Ambig. rate |
|---|---|---|---|---|---|
| Full (14 dec. + 7 metrics) | 0.048 | 67% | 0.902 | 0.98 | 0.04 |
| Own decision + metrics | 0.107 | 26% | 0.910 | 1.05 | 0.10 |
| Metrics only | 0.400 | -176% | 0.936 | 1.54 | 0.55 |

Fig. 2 shows that observed coverage remains similar across feature sets, while accuracy does not. Using weak features, Mondrian achieves 0.94 coverage on the violation class, but this increases predicted set sizes (average size 1.54, with 55% containing both classes), reducing their usefulness for triage. Consequently, the accuracy improvement relies on the single-violation assumption and may not extend to real-world scenarios where multiple violations per image are possible. In contrast, coverage appears stable across different feature qualities within the tested distribution; whether this holds with real multi-attribute data is discussed in Section V.

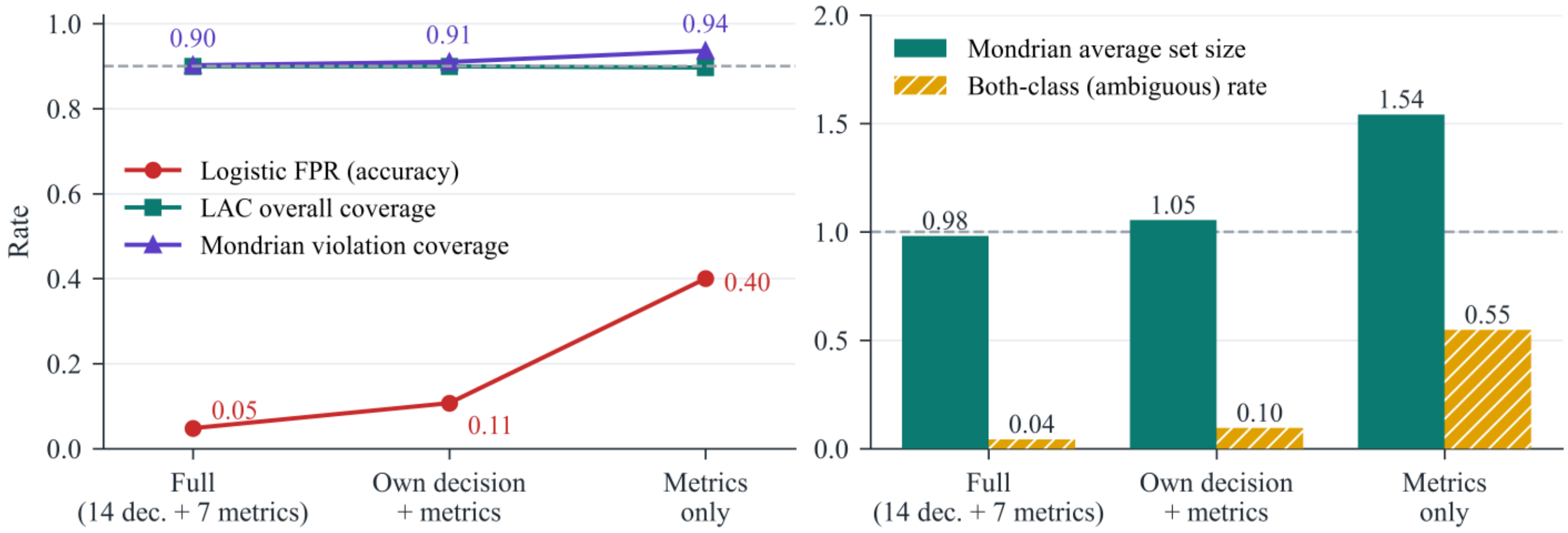

**Figure 2.** Left: as the features are weakened, logistic criterion-level false-positive rate increases while LAC overall coverage and Mondrian violation coverage stay on target. Right: weak features keep coverage only by inflating set size and ambiguity.

### B. MARGINAL CONFORMAL UNDER-COVERS THE VIOLATION CLASS; MONDRIAN BRINGS IT CLOSE TO THE TARGET

Using the exact finite-sample quantile and excluding zoom images (Section III), marginal LAC meets its overall target of approximately 0.90; however, the estimated coverage for each criterion within the violation class is below 0.90 for ten of the fourteen criteria on the primary partition; additionally, the mean values across all partitions (aggregate mean value = 0.81) are below 0.90 for eleven of the fourteen criteria. On the primary partition, the entire 95% cluster-bootstrap confidence interval was below the 0.90 target for three criteria: No posterization 0.39 (0.25 to 0.53); No sunglasses 0.74 (0.61 to 0.86); and In focus 0.77 (0.64 to 0.88). Since there were only 55 test subjects per partition, the clustered confidence intervals were wide. Using the same test, Mondrian's confidence interval was also entirely below 0.90 for two criteria on this partition: No posterization 0.77 (0.66 to 0.87); and Uniform background 0.76 (0.63 to 0.88). We obtained the confidence intervals by resampling only the test subjects while keeping the calibration set and the classifier constant. Thus, a valid method may be below 0.90 on a single partition; Eq. (5) is true on average across calibration sets. Therefore, we evaluate the methods based on the ten-partition means.

Consequently, the ten-partition robustness columns of Table 2 show the individual criterion results: LAC mean scores range from 0.37 (No posterization) to 0.93 and fall below 0.90 on eleven of the fourteen criteria. In contrast, every Mondrian mean score ranges from 0.89 to 0.92. Mondrian thus provides violation-class coverage very close to the nominal target on every criterion and has an aggregate score of 0.91 across all ten partitions, compared with 0.81 for marginal LAC.

**Table 2.** Coverage on the violation class for each criterion. The first columns give the primary-partition estimates with 95% subject-cluster bootstrap CIs and the number of violation images (subjects) in the primary test set; the last two columns give mean ± SD [range] over ten partitions. Marginal LAC under-covers where the classifier is weak; Mondrian's ten-partition means are close to the 0.90 target on every criterion, while single-partition estimates are noisier in both directions.

| ICAO criterion | LAC (95% CI) | Mond. (95% CI) | n viol. (subj.) | LAC, 10 partitions | Mond., 10 partitions |
|---|---|---|---|---|---|
| No posterization | 0.39 (0.25-0.53) | 0.77 (0.66-0.87) | 69 (52) | 0.37 ± 0.07 [0.23-0.45] | 0.89 ± 0.06 [0.77-0.98] |
| No sunglasses | 0.74 (0.61-0.86) | 0.89 (0.79-0.98) | 54 (41) | 0.68 ± 0.11 [0.48-0.83] | 0.91 ± 0.03 [0.87-0.95] |
| In focus | 0.77 (0.64-0.88) | 0.93 (0.87-0.98) | 69 (52) | 0.74 ± 0.07 [0.66-0.89] | 0.89 ± 0.05 [0.77-0.94] |
| No or light makeup | 0.78 (0.64-0.90) | 0.98 (0.93-1.00) | 50 (38) | 0.92 ± 0.06 [0.78-1.00] | 0.91 ± 0.08 [0.73-1.00] |
| Correct exposure | 0.81 (0.72-0.91) | 0.84 (0.75-0.93) | 69 (52) | 0.88 ± 0.04 [0.81-0.94] | 0.89 ± 0.05 [0.77-0.94] |
| Uniform background | 0.85 (0.73-0.94) | 0.76 (0.63-0.88) | 59 (45) | 0.88 ± 0.05 [0.81-0.95] | 0.91 ± 0.06 [0.76-1.00] |
| Uniform face lighting | 0.85 (0.69-0.96) | 0.85 (0.70-0.96) | 27 (25) | 0.87 ± 0.06 [0.79-1.00] | 0.92 ± 0.05 [0.83-1.00] |
| No pixelation | 0.86 (0.77-0.94) | 0.93 (0.85-0.99) | 69 (52) | 0.81 ± 0.03 [0.73-0.86] | 0.90 ± 0.04 [0.83-0.96] |
| Head without covering | 0.86 (0.77-0.95) | 0.82 (0.72-0.91) | 66 (51) | 0.77 ± 0.13 [0.56-0.92] | 0.90 ± 0.06 [0.82-0.98] |
| Neutral expression | 0.87 (0.78-0.94) | 0.86 (0.77-0.93) | 69 (52) | 0.87 ± 0.04 [0.80-0.95] | 0.90 ± 0.04 [0.86-0.95] |
| Frontal pose* | 0.92 (0.77-1.00) | 1.00 (1.00-1.00) | 13 (9) | 0.84 ± 0.16 [0.58-1.00] | 0.92 ± 0.15 [0.60-1.00] |
| Gaze in camera | 0.93 (0.88-0.97) | 0.92 (0.86-0.97) | 137 (52) | 0.90 ± 0.02 [0.87-0.93] | 0.90 ± 0.03 [0.83-0.92] |
| Eyes open | 0.95 (0.91-0.99) | 0.92 (0.87-0.97) | 105 (49) | 0.93 ± 0.04 [0.87-0.98] | 0.92 ± 0.04 [0.84-0.97] |
| Correct saturation | 0.96 (0.90-1.00) | 0.97 (0.93-1.00) | 69 (52) | 0.88 ± 0.08 [0.70-0.98] | 0.91 ± 0.03 [0.87-0.97] |

* Small-sample, reported for completeness only.

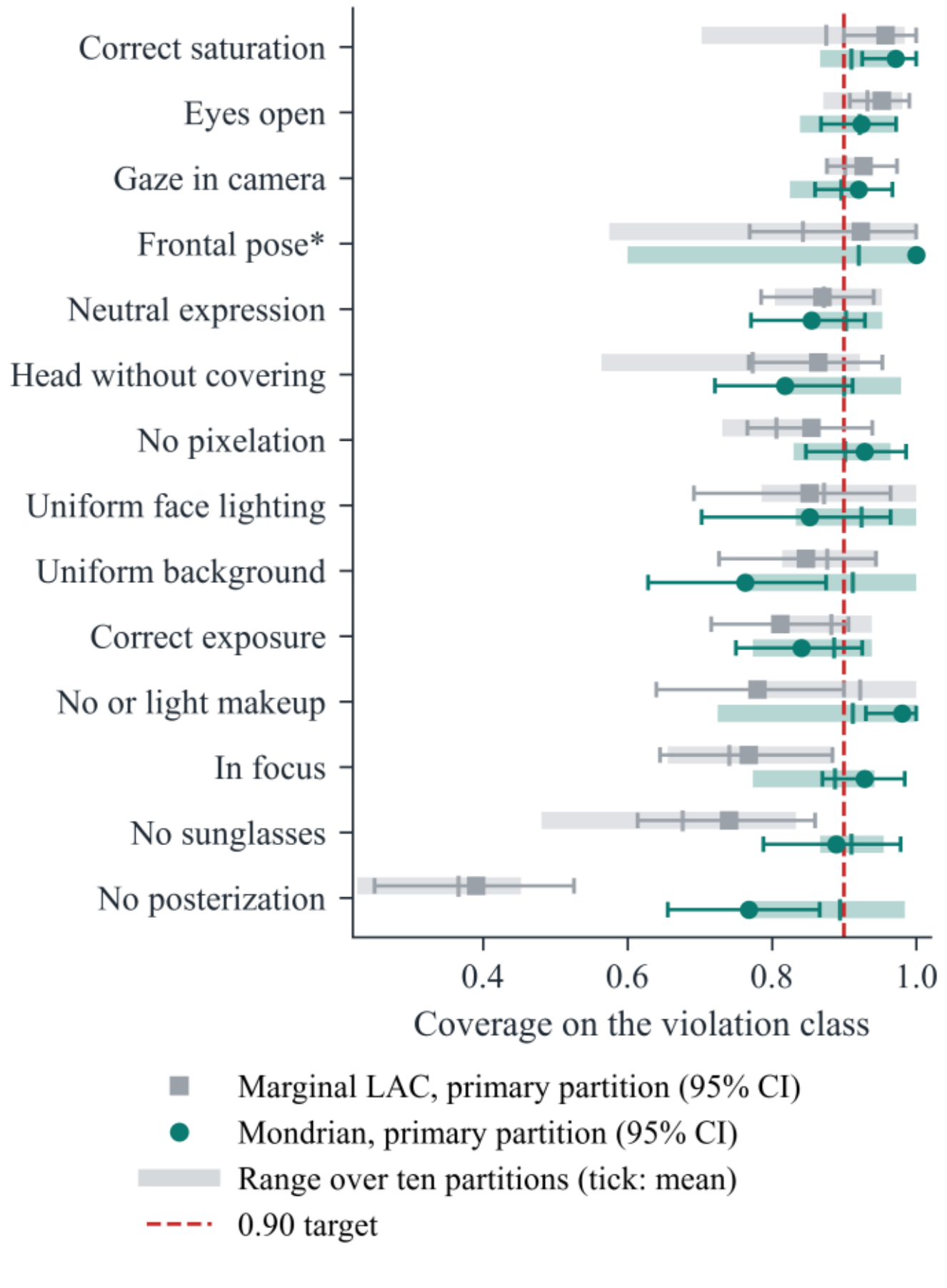

**Figure 3.** Coverage on the violation class for each criterion, marginal LAC (grey) against Mondrian (teal). Markers show the primary partition with 95% subject-cluster bootstrap intervals; the shaded bands show the range over ten partitions, with a tick at the mean. Frontal pose* is flagged as a small sample.

Additionally, there appear to be no measurable adverse effects from Mondrian for criteria where marginal coverage was sufficient (Fig. 3, Table 2). If we remove one degenerate criterion with a very limited number of calibration positives (Frontal pose, with a median of approximately 12 calibration positives across partitions and only 13 test violations from 9 subjects on the primary partition), the average Mondrian violation coverage for the remaining 13 stable criteria is 0.90. We interpret this result as consistent with the intended label-conditional coverage property, i.e., it brings the average violation coverage close to the target level. This does not imply that every criterion will exceed 0.90 on every partition; single-partition estimates are noisy in both directions (on the primary partition, the Mondrian estimates range from 0.76 to 1.00), so we present the clustered intervals together with the ten-partition means and ranges. It is reasonable to expect that Mondrian would improve class-conditional coverage based on its design. What was not obvious in advance is how large the marginal failure is in a real compliance pipeline: down to approximately 0.4 on a criterion whose overall coverage is on target.

### *C. FALSE FLAGS AGAINST DETECTION*

Table 3 shows how the seven strategies performed using the same five partitions. There are three key findings: 1) Most of the reduction in false flags was due to the logistic layer and not to conformal prediction, and most of that was due to TONO's one-violation design (Section IV.A). For example, a simple logistic layer reduced the false positive rate from 0.145 to 0.048, while recall fell only slightly (0.955 to 0.937). 2) Conformal methods did not perform better in terms of accuracy. A plot of recall versus false-positive rate (Fig. 4) showed the conformal operating points on or below the isotonic-calibrated baseline; part of the gap comes from counting a RETAKE as a miss. Therefore, a calibrated classifier with a swept threshold matches or weakly dominates the conformal methods. 3) Once abstention is allowed, the abstention rules tuned on calibration data achieve lower false positive rates than Mondrian at a similar or lower RETAKE rate (Chow 0.027 at 9.6% RETAKE, asymmetric rule 0.015 at 7.3%, Mondrian 0.039 at 10.4%). Additionally, recall among decided cases is similar across all abstaining methods (0.94 to 0.96). Randomised APS is competitive (false positive rate 0.026 at recall 0.89 over all cases). The non-randomised version of APS degrades to trivial two-class sets on this binary classification problem because of a characteristic of that version, not APS itself. Because an abstaining method can improve both the false positive rate (FPR) and recall by directing challenging cases to RETAKE, Table 3 reports recall for all violations and decided cases, along with the RETAKE rate and selective error. The recall values in Table 3 were calculated using the same test partitions and measure a different metric than the violation coverage reported in Table 2. Table 4 displays abstention rules at target values of 0.05 and 0.15. The ranking of the rules remains consistent across target values, and 0.10 is the closest available match point for Mondrian, whose abstention across the entire α range does not fall below approximately 0.10.

**Table 3.** Seven triage strategies under one protocol: mean ± SD over five subject-disjoint partitions. Recall (all) treats abstained violations as missed, while Recall (dec.) is calculated for cases where a decision was rendered.

| Method | FPR | Recall (all) | Recall (dec.) | RETAKE (%) | Viol. acc. (%) | Sel. error | FF redn. |
|---|---|---|---|---|---|---|---|
| Raw BioGaze | 0.145 ± 0.004 | 0.955 ± 0.003 | 0.955 ± 0.003 | 0.0 | 4.5 ± 0.3 | 0.138 | n/a |
| Logistic only | 0.048 ± 0.004 | 0.937 ± 0.012 | 0.937 ± 0.012 | 0.0 | 6.3 ± 1.2 | 0.049 | 67 ± 4% |
| Chow rejection | 0.027 ± 0.003 | 0.800 ± 0.007 | 0.953 ± 0.010 | 9.6 ± 0.7 | 3.9 ± 0.9 | 0.031 | 82 ± 3% |
| Asymmetric two-threshold | 0.015 ± 0.006 | 0.798 ± 0.050 | 0.943 ± 0.016 | 7.3 ± 1.5 | 4.8 ± 1.2 | 0.018 | 90 ± 5% |
| LAC (marginal) | 0.035 ± 0.004 | 0.808 ± 0.016 | 0.942 ± 0.012 | 7.4 ± 0.8 | 4.9 ± 1.0 | 0.039 | 76 ± 4% |
| Mondrian | 0.039 ± 0.004 | 0.878 ± 0.017 | 0.958 ± 0.008 | 10.4 ± 2.8 | 3.9 ± 0.8 | 0.044 | 73 ± 4% |
| Randomised APS | 0.026 ± 0.004 | 0.890 ± 0.017 | 0.959 ± 0.012 | 7.2 ± 0.5 | 3.8 ± 1.1 | 0.029 | 82 ± 3% |

In Table 3, Viol. Acc (%). is the share of true violations that receive a confident compliant decision, the error that matters most for safety; selective error is the error rate among decided cases. The false positive rate (FPR) is calculated using all cases that meet the compliance criteria, including abstained cases; in contrast, Fig. 5 considers only cases with a decision. Thresholds are calibrated exclusively with the calibration data described in Section III.D.

No single method demonstrated overall superiority. The asymmetric rule achieved the lowest FPR and selective error. Mondrian and APS exhibited the highest recall for all cases among abstaining methods. Only the conformal procedures have finite-sample validity results under the corresponding exchangeability assumptions; the present subject-grouped evaluation does not independently establish those assumptions.

**Table 4.** Abstention rules at three calibration targets (same five partitions; Mondrian and APS are target-independent and repeated for reference). RETAKE is in per cent. At each target, the abstention rules achieve a lower FPR than Mondrian at equal or lower RETAKE, except at 0.15, where Chow abstains more. Recall among decided cases is comparable throughout, and what changes with the target is how many violations are abstained rather than detected. At the 0.15 target, the asymmetric rule often falls back to its fixed band (Section III.D), so its pooled abstention reaches only 0.09.

| Method | Target | FPR | Recall (all) | Recall (dec.) | RETAKE | Sel. error |
|---|---|---|---|---|---|---|
| Chow rejection | 0.05 | 0.035 ± 0.003 | 0.876 | 0.948 | 4.9 | 0.038 |
| Chow rejection | 0.10 | 0.027 ± 0.003 | 0.800 | 0.953 | 9.6 | 0.031 |
| Chow rejection | 0.15 | 0.021 ± 0.003 | 0.736 | 0.959 | 14.3 | 0.025 |
| Asymmetric two-threshold | 0.05 | 0.021 ± 0.006 | 0.820 | 0.929 | 5.4 | 0.025 |
| Asymmetric two-threshold | 0.10 | 0.015 ± 0.006 | 0.798 | 0.943 | 7.3 | 0.018 |
| Asymmetric two-threshold | 0.15 (0.09 reached) | 0.016 ± 0.004 | 0.785 | 0.950 | 9.0 | 0.019 |
| Mondrian (α = 0.10) | n/a | 0.039 ± 0.004 | 0.878 | 0.958 | 10.4 | 0.044 |
| Randomised APS (α = 0.10) | n/a | 0.026 ± 0.004 | 0.890 | 0.959 | 7.2 | 0.029 |

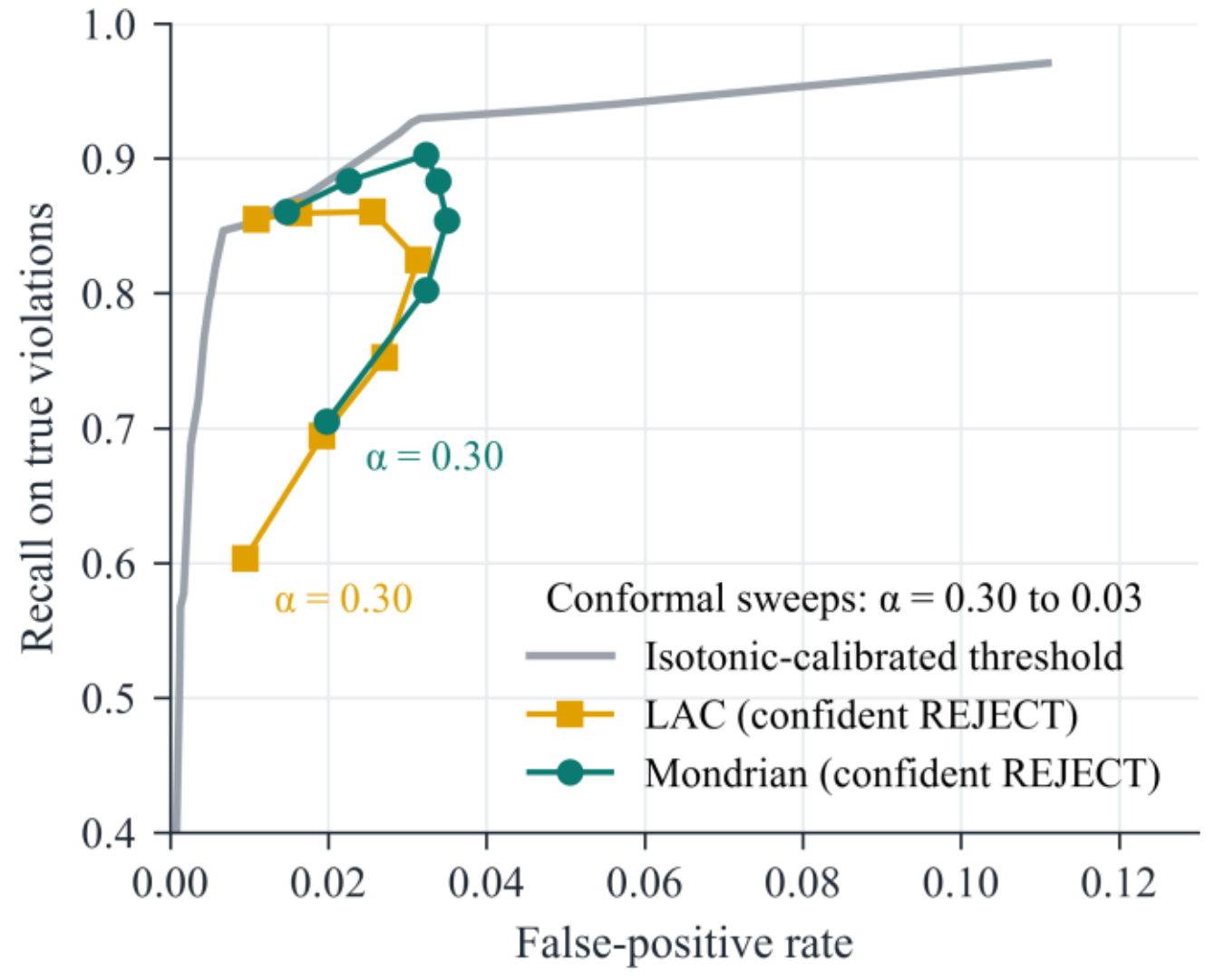

**Figure 4.** Recall versus false-positive rate (primary partition). RETAKE counts as a missed detection; conformal operating points lie on or below the isotonic-calibrated baseline.

**Tuning and set types.** Chow's τ and the asymmetric pair ($t_{lo}$, $t_{hi}$) are set on the calibration split to a fixed abstention target of 0.10, and the test split is never used. On test, Mondrian's prediction sets are 90% singleton, 4% ambiguous and 6% empty (LAC 92.6/1.0/6.4; APS 93/7/0). Coverage on the violation class over the same partitions is 0.82 (LAC), 0.90 (Mondrian) and 0.96 (APS), so APS over-covers the violation class at the 0.90 level. APS offers a good trade-off here, but it provides no class-conditional guarantee, so on other data it could under-cover the violation class; that is why it is not our class-conditional recommendation.

Fig. 5 shows how each abstention method performs across operating points. Each point plots the false positive rate (FPR) of the decided cases against the abstention rate (RETAKE). Conformal methods are shown as points across a range of α values. As α decreases, the abstention rate of conformal methods decreases because empty sets disappear.

However, as α continues to decrease and sets become ambiguous, the abstention rate increases. Across the operating points evaluated in Fig. 5, calibration-tuned abstention rules attain an FPR no higher than the displayed conformal points at similar RETAKE rates. For these reported operating points, tuned rules are competitive on FPR and abstention; no method dominates once the class-conditional guarantee is also counted.

Thus, a conventional abstention rule provides a similar or better empirical operating point. It does not, however, provide a controllable class-conditional guarantee. On a binary criterion, conformal methods are threshold rules. Marginal LAC is, on a criterion-by-criterion basis, Chow's rule with τ equal to max(q, 1 − q) for the single calibration quantile q (when pooled over 14 criteria, the LAC points in Fig. 5 deviate from the Chow curve only because LAC sets q for each criterion, whereas the Chow sweep uses one common τ). At the same time, Mondrian is a class-conditional rule with two thresholds, i.e., the calibration quantiles of the nonconformity score within each class.

Therefore, Table 3 compares two families of rules: a symmetric confidence band and an asymmetric band with two thresholds, both used twice: once with thresholds fixed as quantiles at the nominal level and once with thresholds searched against an abstention or cost objective. The choice of quantiles determines Mondrian's properties. Its coverage on the violation class is determined directly by the nominal level $1 - \alpha$ and holds in finite samples under exchangeability, without searching for thresholds against a cost objective. Fig. 5 (right) shows the achieved violation coverage relative to the nominal target across the sweep; it sits 0.003 to 0.011 below nominal at the four levels from 0.70 to 0.90 and at or slightly above nominal at tighter levels. With five partitions, the standard error of this mean is about 0.006, so these shortfalls are within sampling noise. The shortfall is consistent with the exchangeability caveat of Section IV.G. By contrast, the violation coverage of the abstention rules is an uncontrolled by-product of tuning a band. Moreover, marginal LAC under-covers the violation class (0.82 at the 0.90 level), although it meets its overall target. Consequently, Mondrian brings a formal procedure for selecting the two thresholds. This procedure has a finite-sample validity property that a tuned rule lacks, and our results show that marginal quantiles fail on the minority class. It is not a win on FPR or recall.

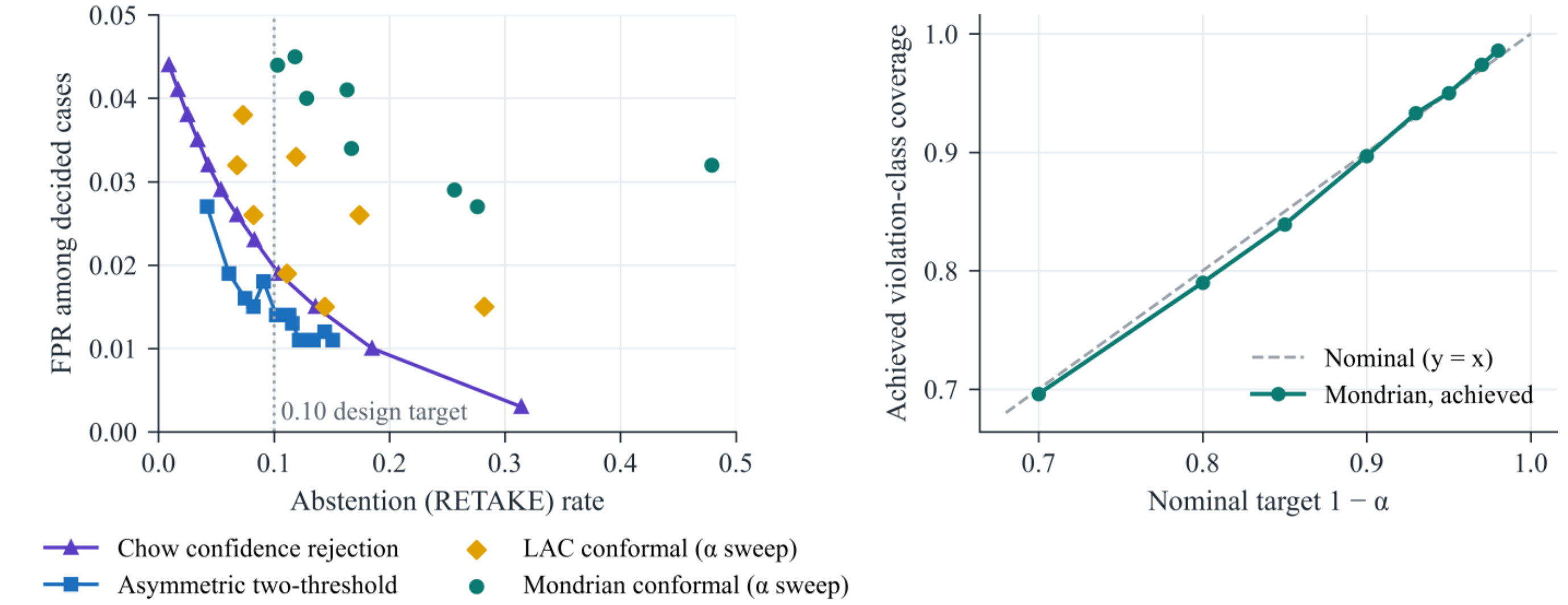

**Figure 5.** Left: risk against abstention (five partitions pooled), false positive rate among decided cases against RETAKE rate. The abstention rules tuned on calibration data (lines) reach an FPR lower than or equal to the conformal methods (points, α sweep) at comparable displayed abstention rates. Right: Mondrian's achieved coverage on the violation class tracks the nominal target $1-\alpha$, which the abstention rules do not do.

### D. EXPLORATORY IMAGE-LEVEL TRIAGE ON TONO

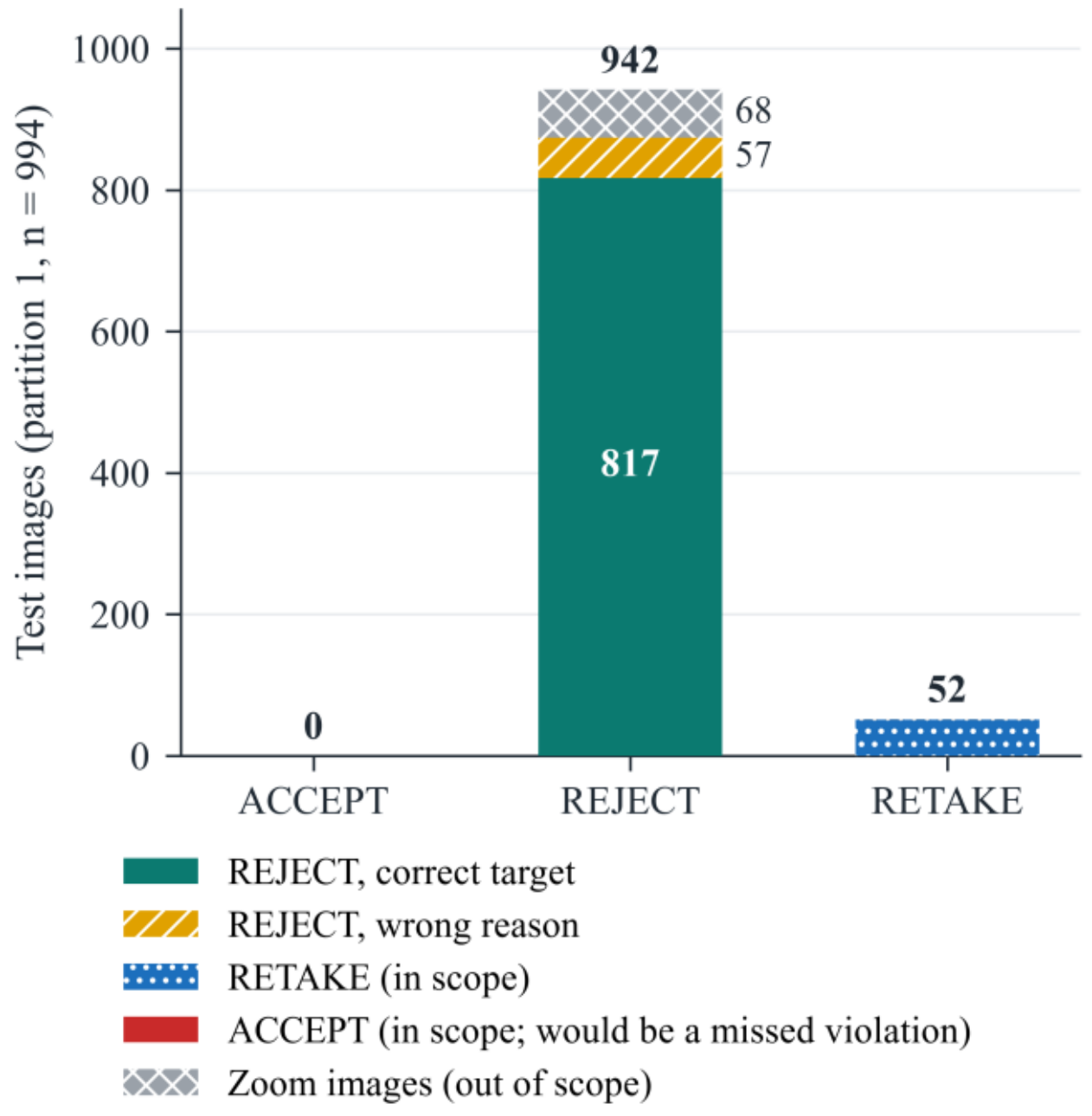

**Figure 6.** Triage at the image level under Mondrian (partition 1), with REJECT split into correct target, wrong reason and zoom images out of scope. No violation image within scope is wrongly accepted; ACCEPT is empty because TONO contains no genuinely compliant images within scope.

On the first partition of the test set of 994 images (Fig. 6), combining the Mondrian sets for each criterion results in 0 ACCEPT, 942 REJECT and 52 RETAKE decisions. The first partition of the test set consists of 925 images that have a violation within scope and 69 zoom images that are out of scope (Section III.B). Because each image with a violation has exactly one violation, we can estimate image-level specificity only using the 69 zoom images, which meet all 14 evaluated criteria. None of the 69 zoom images was accepted (0/69; 68 were rejected and one was sent to RETAKE). We excluded zoom images from training and calibration because they fall outside the distribution of the head size metric. As a result, the specificity estimate is limited. Nevertheless, this is the only way TONO allows us to measure specificity, and on this proxy, image-level specificity is 0/69. Because a naive REJECT count conflates correct and incorrect rejections, we evaluated the 925 violation images based on whether the identified criterion matched the actual violation. There were 817 REJECT decisions on the correct target (88.3%; the real violation was identified and flagged confidently), 57 REJECT decisions for the wrong reason (6.2%; the image was rejected based on a different criterion than the one violated, so the actual violation was not flagged), 51 RETAKE decisions (5.5%) and 0 ACCEPT decisions.

Overall, the triage correctly flags the criteria in most instances; the 6.2% of violation images rejected for the wrong reason represent a failure-to-report mode, and because the ACCEPT class is empty, the zero accepted violations provide no information about safety. We are not asserting a general safety property. All TONO images contain one violation; therefore, there are virtually no truly compliant images, the ACCEPT class is empty, and specificity at the image level (the risk of incorrectly rejecting a compliant photo) cannot be assessed here. To evaluate specificity, a real-world dataset with multiple attributes per image and a true, in-distribution ACCEPT class would be required.

### *E. JOINT COVERAGE AND THE MULTIPLE-COMPARISON PROBLEM*

Each per-criterion guarantee applies to only one criterion at a time. Without correcting for multiple comparisons, the empirical joint coverage of all 14 prediction sets is only 0.27 (mean over five partitions). This aligns with what we would expect from multiple comparisons. We can apply a Bonferroni correction [33] (target 0.993 per criterion) to increase total empirical joint coverage to 0.91 for both marginal and Mondrian conformal; however, this comes at a cost. For marginal LAC, approximately 96% of images are sent to RETAKE. For Mondrian, a level of 0.993 requires at least 139 calibration positives per criterion; the largest criterion (Gaze in camera) has 93 to 143 across the five partitions (median 114), so the corrected violation threshold is almost always infinite, the violation class is almost always included, and every image goes to RETAKE (Fig. 7). The marginal method keeps a finite threshold because it pools both classes (about 820 calibration images on average). It reaches 0.91 joint coverage by sending 96% of the images to RETAKE. Here, an image counts as RETAKE if any criterion set is ambiguous or empty, which is stricter than the triage rule of Section IV.D.

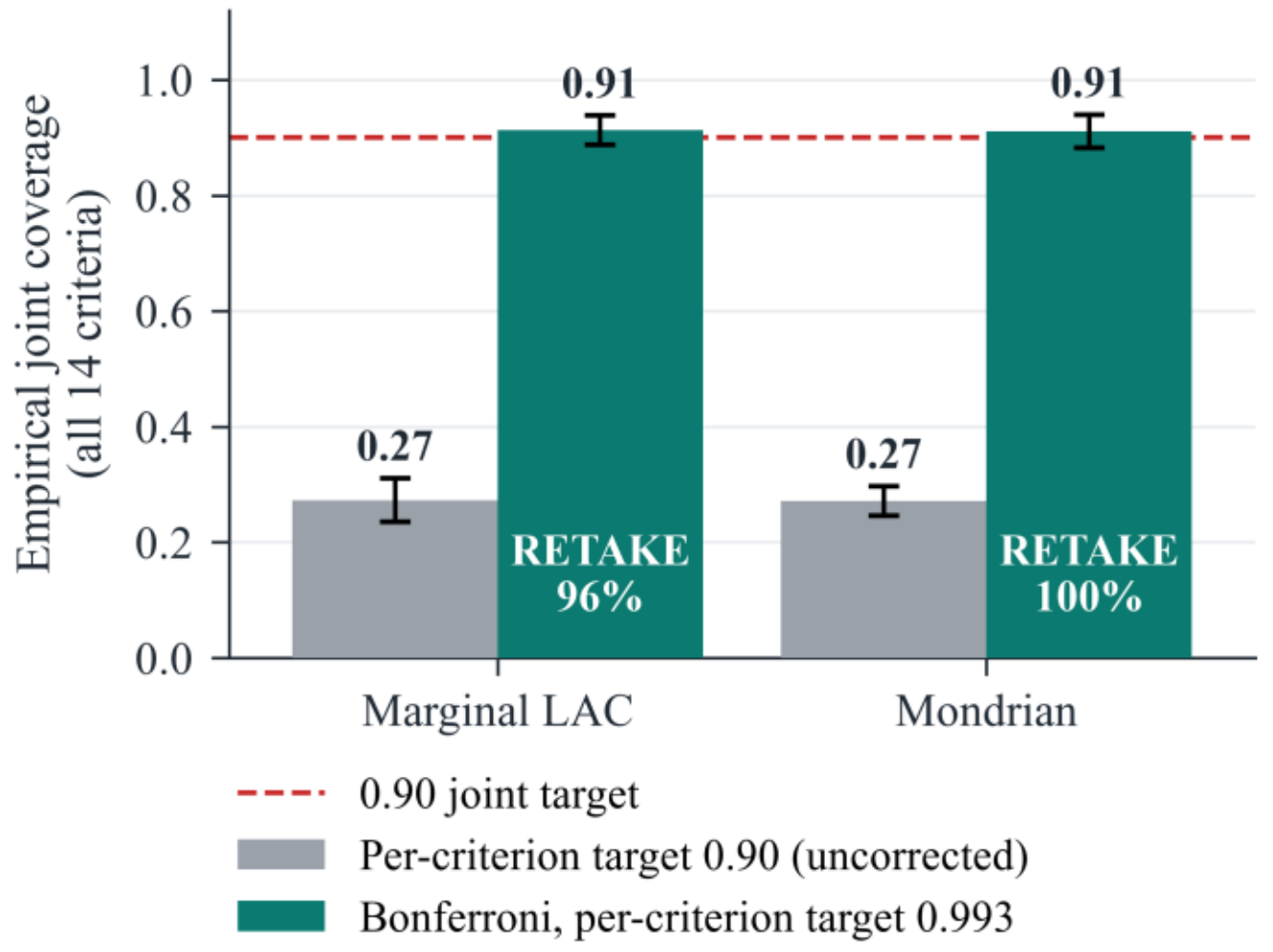

**Figure 7.** Naive joint coverage over 14 criteria collapses to about 0.27. A Bonferroni correction brings empirical joint coverage to about 0.91. Still, it sends 96% (LAC) or all (Mondrian, whose corrected violation threshold is almost always infinite at this calibration size) of the images to RETAKE.

The error that is most important for safety does not need a multiple-comparison correction. An image receives ACCEPT only if all 14 sets are {compliant}, so a violating image is accepted only if the set for its violated criterion misses the violation. Under Mondrian, this has probability at most α for each criterion, on average over calibration and test draws. A correction is needed only for the joint coverage of all 14 labels.

Therefore, while a Bonferroni guarantee for an image is feasible, it is not practical at this scale, and future research will explore a tighter simultaneous guarantee.

### *F. ROBUSTNESS*

The results were consistent across all validation tests. The overall marginal coverage (coverage averaged over all test images, regardless of class) closely matched the nominal target values for the three levels of coverage tested (nominal targets were 0.95, 0.90, and 0.80; actual empirical coverages were 0.951, 0.900, and 0.803). Mondrian's violation coverage tracks the nominal level across the α sweep (Fig. 5, right). It stays close to 0.90 in each of the ten partitions (Table 2, Fig. 3). On average, violation coverage with Mondrian ranged from 0.88 to 0.92, whereas without Mondrian it ranged from 0.78 to 0.84. Randomised APS shows a similar pattern across partitions and across both independent randomisation streams used in our two analysis scripts. (Recall that randomised APS provided average recall of 0.887 and 0.890; the average size of the set of selected items was approximately 1.07). We made no variations to the split ratio.

### *G. EXCHANGEABILITY AND SENSITIVITY TO CLUSTERING*

Split conformal prediction requires calibration and test samples to be exchangeable. By making the subjects disjoint, we remove identity leakage, but this does not create exchangeability at the image level. This is because a subject provides multiple correlated images. Throughout the remainder of the paper, we take the position that the guarantee holds at the image level under the assumption of exchangeability; we use disjoint subjects to protect against identity leakage, not to prove exchangeability; and we empirically assess the impact of dependence within subjects. The data have the following structure. Each of the five partitions of Table 3 contains 165 training subjects, 55 calibration subjects and 55 test subjects. The calibration portion of the partition contains between 760 and 1020 images (median 14-15 images per subject). The test portion contains between 780 and 1080 images (median 14-15 images per subject). The subject contributing the most images to the calibration set contributed, on average, about 8% of the calibration set (maximum 14%), while the subject contributing the most images to the test set contributed, on average, about 10% of the test set (maximum 16%). No subject contributed more than 16% of either the calibration or the test set. To evaluate whether within-subject dependence influenced the results, we separated clustering effects from sample-size effects. When we limit each subject to three calibration images, the calibration set has no more than 165 images (3 per calibration subject) and yields a Mondrian violation coverage of 0.92 (SD 0.02 across partitions). When we randomly subsample images of the same size without regard to subjects, we get 0.92 (SD 0.02). Since both results are similar, we found no evidence of clustering at equal n. The change from 0.90 to 0.92 reflects the conservatism of the exact quantile for small n, not a subject effect. Bootstrapping over subjects from the full calibration set (resampling subjects with replacement and including multiplicity) yields a mean coverage of 0.89, with a 5th-95th percentile range of 0.86 to 0.91, which includes the 0.90 target.

The standard deviation of coverage across test subjects is approximately 0.12, reflecting the heterogeneity a subject-level guarantee must address.

This evaluation has limited statistical power because both the capped and random subsamples contain approximately three images per subject. Consequently, the difference in clustering between the two groups is minimal because of how they were constructed. A more direct diagnostic is the distribution of Mondrian violation coverage across the five partitions. The standard deviation is 0.013 across the five partitions of Table 3, and 0.014 across the ten partitions of Section IV.B, against a binomial value of approximately 0.010 expected under independence at roughly 850 violation evaluations per partition. The excess is modest, and it points in the same direction as the subject bootstrap: a mild within-subject effect that widens the uncertainty slightly without moving the means. Within the constraints of these assessments, the principal finding does not appear to be an artefact of within-subject dependence. A formal clustered conformal guarantee, or a subject-level one, remains an area for future research.

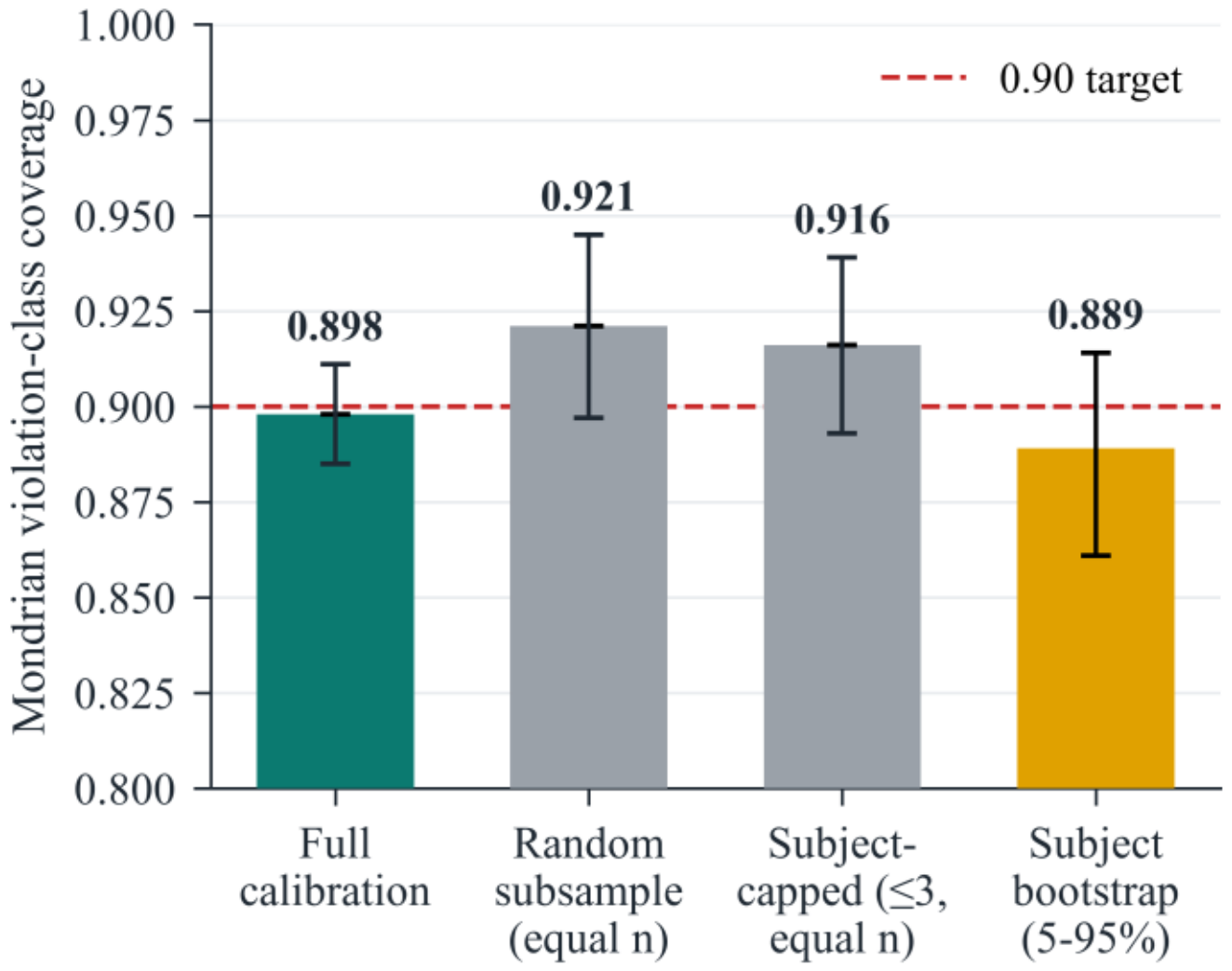

**Figure 8.** Sensitivity of Mondrian coverage on the violation class to clustering. Subsamples capped by subject and random subsamples of equal n agree; this check has limited power, and the between-partition SD (Section IV.G) indicates at most a mild within-subject effect; the range of the bootstrap over subjects contains the 0.90 target.

## V. DISCUSSION

**Which pipeline suits which objective?** Based on our findings, we identified two straightforward operational interpretations. If the primary goal is to minimise false positives without an abstention mechanism, most of the reduction comes from applying a logistic layer to the analyser's output, so an abstention mechanism is unnecessary. If, however, there is a budget for retakes and the optimal operating point is to be found through empirical testing, a tuned abstention rule, such as the asymmetric rule used in this study, delivers the lowest false-positive rate while maintaining a similar or lower RETAKE rate. This applies to the asymmetric rule as described in this study, including the fixed-band fallback (Section III.D). Mondrian conformal is justified when there is a binding requirement for a documented and auditable statement regarding the miss rate on true violations. The coverage level for the violation class is set as a policy parameter before implementation and, under exchangeability, holds in finite samples on average. The auditor requires no threshold searching. However, this comes at the expense of a 1-3 percentage point increase in the false-positive rate relative to a similar RETAKE rate. Determining whether this additional cost is warranted is the responsibility of the issuing authority, not this paper.

**What does α control?** For a label-conditional Mondrian procedure, α limits the miscoverage rate within each class, including the violation class, but it does not control the RETAKE rate, which is determined by the prediction sets. Our research found that Mondrian conformal's abstention rate never fell below approximately 10% across the tested values of α and increased at both ends of the sweep. Therefore, simply changing α cannot impose a hard limit on retakes; this would require either a tuned abstention rule or a budget mechanism layered on top of the conformal layer. By contrast, the tuned rules used in this work do not provide a certified miss rate: violation coverage is a by-product of their thresholds, and certifying it would require an additional step, such as conformal risk control [15] on held-out data.

**Robustness within the evaluated distribution versus external validity.** Within the evaluated distribution, the observed coverage property behaved as described by theory. It stayed on target for all three feature sets in the ablation and remained consistent across ten different partitions. The clustering diagnostics in Section IV.G indicate at most a mild effect within subjects.

However, none of these findings establishes validity outside of TONO. Our dataset is artificially generated, contains exactly one violation per image, and has no truly compliant in-scope images. Consequently, we have not tested the ACCEPT decision, image-level specificity, or behaviour under multiple concurrent violations. As such, we claim narrowly that the coverage property is robust to feature quality within the evaluated distribution, and we have yet to determine its validity on real, multi-attribute data.

The next logical step for validating the pipeline is to use the DFIC dataset [24]. DFIC has 58,633 images of 1,016 subjects, annotated against 26 requirements. In addition, DFIC includes both compliant and non-compliant images of actual subjects; TONO does not.

By running the same pipeline on DFIC, we can evaluate the validity of our claims and determine whether our methods apply to real-world scenarios. Specifically, we would like to know whether marginal conformal still does not adequately address the violation class in real-world data, whether Mondrian continues to provide us with sets that are large enough to be useful while keeping them close to the target, whether the accuracy artefact described in Section IV.A disappears when multiple violations are present in one image, as predicted, and whether the ACCEPT/REJECT decision can now be measured at the image level using controlled specificities for the first time.

In addition to testing our claims, running the pipeline on DFIC will let us calibrate a RETAKE region based on the actual costs of missing a violation and performing a retake. We believe this external validation is the most important test of the paper's main claim.

### A. LIMITATIONS

- **Logistic layer versus conformal.** Most of the decrease in false positives comes from the logistic layer alone; conformal adds both a distribution-free guarantee and, via Mondrian, a class-conditional guarantee, not increased accuracy.

- **One violation per image.** TONO's design lets the classifier use signals from other criteria; thus, the observed results are upper bounds based on the dataset used and are unlikely to apply to images with multiple violations. The coverage property does not depend on the features within the evaluated distribution; its validity on real data remains untested.
- **Empty ACCEPT class.** No truly compliant images are in scope for TONO, so we cannot validate image-level specificity or an ACCEPT decision in distribution; the only proxy, 69 out-of-scope zoom images, yielded no ACCEPT decisions (68 REJECT, 1 RETAKE).
- **Guarantee at the subject level.** We report an empirical diagnostic at the subject level, not a formal clustered conformal guarantee, and the diagnostic may fail to detect a clustering effect smaller than its sampling noise.
- **Subject-level validity.** The current subject-cluster bootstrap and capped-subsample analyses are diagnostics, not a finite-sample hierarchical conformal guarantee. HCP provides a subject-weighted calibration construction for predicting an observation from a previously unseen exchangeable group [26]. Combining subject-aware validity with class-conditional validity for rare ICAO violations requires a separately specified method and evaluation.
- **Validity versus usefulness.** Mondrian's benefit lies in validity rather than in accuracy or false positives: an abstention rule tuned on calibration data yields a lower false-positive rate at a comparable retake rate. Whether the formal guarantee is worth the cost is an operational decision informed by this study but not resolved.
- **Single-partition intervals.** The primary intervals shown in Table 2 were calculated using a subject-cluster bootstrap applied to a single partition of the data. While these intervals account for correlations between images of the same test subject, they do not capture the variability that arises when the data are re-partitioned (as shown in the ten-partition columns). Because each partition contains 55 test subjects, some intervals are wide, which is common in this type of clustered design.
- **Joint coverage.** Guarantees for individual criteria do not combine into a single guarantee for the image without applying a conservative correction for multiple comparisons.
- **Criteria with low sample sizes.** Due to small sample sizes for rare criteria (e.g., Frontal pose), coverage estimates are highly variable. We thus flag these criteria rather than rely on them.

The external validation on DFIC discussed above addresses dataset-bound limitations; the methodological ones, namely a formal clustered guarantee and a non-vacuous simultaneous guarantee, remain open problems.

## VI. CONCLUSION

We studied a conformal prediction layer over a rule-based analyser for 14 ICAO face-image compliance criteria, which provides a distribution-free marginal coverage guarantee if exchangeability holds. This guarantee holds on average across images, not for every image. No subject was common to both the calibration and test sets. One of the main empirical findings of this study is that the marginal method failed substantially for the minority violation class (violation-class coverage as low as approximately 0.4 on a criterion where overall coverage is on target).

However, label-conditional Mondrian conformal methods bring violation-class coverage closer to the nominal target, with every ten-partition mean between 0.89 and 0.92, consistent with label-conditional validity. Furthermore, a feature ablation study shows that the coverage properties of the models developed are robust to the feature set used in the evaluation distribution. However, model performance depends on whether each image contains only one violation, so we must also evaluate these models on real-world data with multiple attributes. We also evaluated seven triage methods using thresholds established from calibration data. Most performance gains came from the logistic layer, while traditional abstention rules achieved similar or better empirical operating points. Mondrian is distinct from other methods because it provides a formal class-conditional coverage guarantee. Finally, we account for the multiple-comparison issue involved in making decisions across 14 criteria and limit our claims to what the available data support.

## ACKNOWLEDGMENT

The authors thank the BioLab group at the University of Bologna for making the TONO dataset and the BioGaze framework publicly available. We used Grammarly to check and edit the language of the manuscript. We used Claude (Anthropic) to help draft and edit parts of the text and to help write the analysis code, and we checked the figures and reported numbers with its help. All text, code and results were reviewed and revised by the authors, who take full responsibility for the content of this paper.

## DATA AVAILABILITY

The analysis code, the result files and the scripts for Figs. 2 to 8 are available at GitHub https://github.com/mvy0502/icao-conformal-repo and archived on Zenodo https://doi.org/10.5281/zenodo.2367434.

The per-image BioGaze outputs on TONO are released in the same repository under CC BY-NC 4.0, the licence of the TONO dataset [22]; the TONO images themselves are available from their authors.

## References

[1] International Civil Aviation Organization, *Doc 9303, Machine Readable Travel Documents, Part 9: Deployment of Biometric Identification and Electronic Storage of Data in MRTDs*, eighth ed., ICAO, Montreal, 2021, [Online]. Available at: https://www.icao.int/publications/doc-series/doc-9303

[2] ISO/IEC 39794-5:2019, *Information technology, Extensible biometric data interchange formats, Part 5: Face image data*, first ed., International Organization for Standardization, 2019, [Online]. Available at: https://www.iso.org/standard/72156.html

[3] O. Elatfi, N. Di Domenico, G. Borghi, A. Franco, D. Maltoni, "BioGaze: a framework for evaluating the photographic requirements of the ISO/IEC 39794-5 standard," *Proceedings of the 2025 IEEE 19th International Conference on Automatic Face and Gesture Recognition (FG)*, Tampa/Clearwater, FL, USA, May 26-30, 2025, pp. 1-10, https://doi.org/10.1109/FG61629.2025.11099423.

[4] M. Sadinle, J. Lei, L. Wasserman, "Least ambiguous set-valued classifiers with bounded error levels," *Journal of the American Statistical Association*, vol. 114, issue 525, pp. 223-234, 2019, https://doi.org/10.1080/01621459.2017.1395341.

[5] Y. Romano, M. Sesia, E.J. Candès, "Classification with valid and adaptive coverage," *Advances in Neural Information Processing Systems*, vol. 33, 2020, pp. 3581-3591.

[6] V. Vovk, A. Gammerman, G. Shafer, *Algorithmic Learning in a Random World*, second ed., Springer, Cham, 2022, https://doi.org/10.1007/978-3-031-06649-8.

[7] G. Shafer, V. Vovk, "A tutorial on conformal prediction," *Journal of Machine Learning Research*, vol. 9, issue 12, pp. 371-421, 2008.

[8] A.N. Angelopoulos, S. Bates, "Conformal prediction: a gentle introduction," *Foundations and Trends in Machine Learning*, vol. 16, issue 4, pp. 494-591, 2023, https://doi.org/10.1561/2200000101.

[9] V. Vovk, "Conditional validity of inductive conformal predictors," *Machine Learning*, vol. 92, issue 2-3, pp. 349-376, 2013, https://doi.org/10.1007/s10994-013-5355-6.

[10] T. Ding, A.N. Angelopoulos, S. Bates, M.I. Jordan, R.J. Tibshirani, "Class-conditional conformal prediction with many classes," *Advances in Neural Information Processing Systems*, vol. 36, 2023, pp. 64555-64576, https://doi.org/10.52202/075280-2817.

[11] C.K. Chow, "On optimum recognition error and reject tradeoff," *IEEE Transactions on Information Theory*, vol. 16, issue 1, pp. 41-46, 1970, https://doi.org/10.1109/TIT.1970.1054406.

[12] R. El-Yaniv, Y. Wiener, "On the foundations of noise-free selective classification," *Journal of Machine Learning Research*, vol. 11, issue 53, pp. 1605-1641, 2010.

[13] Y. Geifman, R. El-Yaniv, "Selective classification for deep neural networks," *Advances in Neural Information Processing Systems*, vol. 30, 2017, pp. 4878-4887.

[14] Y. Geifman, R. El-Yaniv, "SelectiveNet: a deep neural network with an integrated reject option," *Proceedings of the 36th International Conference on Machine Learning*, Proceedings of Machine Learning Research, vol. 97, 2019, pp. 2151-2159.

[15] A.N. Angelopoulos, S. Bates, A. Fisch, L. Lei, T. Schuster, "Conformal risk control," *Proceedings of the International Conference on Learning Representations (ICLR)*, 2024, https://doi.org/10.48550/arXiv.2208.02814.

[16] S. Bates, A. Angelopoulos, L. Lei, J. Malik, M.I. Jordan, "Distribution-free, risk-controlling prediction sets," *Journal of the ACM*, vol. 68, issue 6, pp. 1-34, 2021, https://doi.org/10.1145/3478535.

[17] M. Ferrara, A. Franco, D. Maio, D. Maltoni, "Face image conformance to ISO/ICAO standards in machine readable travel documents," *IEEE Transactions on Information Forensics and Security*, vol. 7, issue 4, pp. 1204-1213, 2012, https://doi.org/10.1109/TIFS.2012.2198643.

[18] T. Schlett, C. Rathgeb, O. Henniger, J. Galbally, J. Fierrez, C. Busch, "Face image quality assessment: a literature survey," *ACM Computing Surveys*, vol. 54, issue 10s, pp. 1-49, 2022, https://doi.org/10.1145/3507901.

[19] P. Terhörst, M. Huber, N. Damer, F. Kirchbuchner, K. Raja, A. Kuijper, "Pixel-level face image quality assessment for explainable face recognition," *IEEE Transactions on Biometrics, Behavior, and Identity Science*, vol. 5, issue 2, pp. 288-297, 2023, https://doi.org/10.1109/TBIOM.2023.3263186.

[20] Federal Office for Information Security (BSI), *Open Source Face Image Quality (OFIQ) 1.0, reference implementation of ISO/IEC 29794-5*, 2024, [Online]. Available at: https://www.bsi.bund.de/dok/OFIQ1-e

[21] N. Di Domenico, G. Borghi, A. Franco, D. Maltoni, "ONOT: a high-quality ICAO-compliant synthetic mugshot dataset," *Proceedings of the 2024 IEEE 18th International Conference on Automatic Face and Gesture Recognition (FG)*, Istanbul, Turkiye, May 27-31, 2024, pp. 1-10, https://doi.org/10.1109/FG59268.2024.10581986.

[22] G. Borghi, A. Franco, N. Di Domenico, D. Maltoni, "TONO: A synthetic dataset for face image compliance to ISO/ICAO standard," in: A. Del Bue, C. Canton, J. Pont-Tuset, T. Tommasi (Eds.), *Computer Vision - ECCV 2024 Workshops, Part XX*, Lecture Notes in Computer Science, vol. 15642, Springer, Cham, 2025, pp. 1-18, https://doi.org/10.1007/978-3-031-91907-7_1.

[23] N. Di Domenico, G. Borghi, A. Franco, D. Maltoni, "Towards zero-shot ISO/ICAO face compliance verification via CLIP-IQA and natural language prompting," *Proceedings of the 2025 IEEE International Joint Conference on Biometrics (IJCB)*, Osaka, Japan, September 8-11, 2025, pp. 1-9, https://doi.org/10.1109/IJCB65343.2025.11410834.

[24] N. Gonçalves, D. Nunes, C. Guerra, J. Marcos, "DFIC: towards a balanced facial image dataset for automatic ICAO compliance verification," *arXiv preprint*, arXiv:2602.10985, 2026, https://doi.org/10.48550/arXiv.2602.10985.

[25] A.G.A. Silva, H.M. Gomes, L.V. Batista, "A collaborative deep multitask learning network for face image compliance to ISO/IEC 19794-5 standard," *Expert Systems with Applications*, vol. 198, article 116756, 2022, https://doi.org/10.1016/j.eswa.2022.116756.

[26] Y. Lee, R. Foygel Barber, R. Willett, "Distribution-free inference with hierarchical data," *arXiv preprint*, arXiv:2306.06342, 2023, https://doi.org/10.48550/arXiv.2306.06342.

[27] J. Hallberg Szabadváry, T. Löfström, U. Johansson, C. Sönströd, E. Ahlberg, L. Carlsson, "Classification with reject option: Distribution-free error guarantees via conformal prediction," *Machine Learning with Applications*, vol. 20, article 100664, 2025, https://doi.org/10.1016/j.mlwa.2025.100664.

[28] N. Fatima, M. Ahmed, "Picture the suspect: Conformal soft-biometric triage across faces and pedestrians," *Computer Vision and Image Understanding*, vol. 271, article 104870, 2026, https://doi.org/10.1016/j.cviu.2026.104870.

[29] H. Papadopoulos, K. Proedrou, V. Vovk, A. Gammerman, "Inductive confidence machines for regression," in: T. Elomaa, H. Mannila, H. Toivonen (Eds.), *Machine Learning: ECML 2002*, Lecture Notes in Computer Science, vol. 2430, Springer, Berlin, Heidelberg, 2002, pp. 345-356, https://doi.org/10.1007/3-540-36755-1_29.

[30] C.A. Field, A.H. Welsh, "Bootstrapping clustered data," *Journal of the Royal Statistical Society: Series B (Statistical Methodology)*, vol. 69, issue 3, pp. 369-390, 2007, https://doi.org/10.1111/j.1467-9868.2007.00593.x.

[31] V. Taquet, V. Blot, T. Morzadec, L. Lacombe, N. Brunel, "MAPIE: an open-source library for distribution-free uncertainty quantification," *arXiv preprint*, arXiv:2207.12274, 2022, https://doi.org/10.48550/arXiv.2207.12274.

[32] B. Zadrozny, C. Elkan, "Transforming classifier scores into accurate multiclass probability estimates," *Proceedings of the Eighth ACM SIGKDD International Conference on Knowledge Discovery and Data Mining*, ACM, New York, 2002, pp. 694-699, https://doi.org/10.1145/775047.775151.

[33] O.J. Dunn, "Multiple comparisons among means," *Journal of the American Statistical Association*, vol. 56, issue 293, pp. 52-64, 1961, https://doi.org/10.1080/01621459.1961.10482090.